\documentclass{article}
\usepackage{spconf, amsmath,graphicx}
\usepackage{flushend}
\pdfoutput=1
\usepackage{hyperref}
\usepackage{epstopdf}
\title{Resource Aware Design of a Deep Convolutional-Recurrent Neural Network for Speech Recognition Through Audio-Visual Sensor Fusion}

\name{Matthijs Van keirsbilck \qquad Bert Moons \qquad Marian Verhelst}
\address{MICAS, Department of Electrical Engineering (ESAT), KU Leuven, Leuven, Belgium}
\hypersetup{pdfauthor={Matthijs Van keirsbilck},pdftitle={Resource Aware Design of a Deep Convolutional-Recurrent Neural Network for Speech Recognition Through Audio-Visual Sensor Fusion}}

\begin{document}
\setlength{\textfloatsep}{0.3cm}  
\ninept	

\maketitle
\begin{abstract}
	Today's Automatic Speech Recognition systems only rely on acoustic signals and often don't perform well under noisy conditions. Performing multi-modal speech recognition - processing acoustic speech signals and lip-reading video simultaneously - significantly enhances the performance of such systems, especially in noisy environments.
	This work presents the design of such an audio-visual system for Automated Speech Recognition, taking memory and computation requirements into account. First, a Long-Short-Term-Memory neural network for acoustic speech recognition is designed. Second, Convolutional Neural Networks are used to model lip-reading features. These are combined with an LSTM network to model temporal dependencies and perform automatic lip-reading on video. Finally, acoustic-speech and visual lip-reading networks are combined to process acoustic and visual features simultaneously. An attention mechanism ensures performance of the model in noisy environments. This system is evaluated on the TCD-TIMIT 'lipspeaker' dataset for audio-visual phoneme recognition with clean audio and with additive white noise at an SNR of 0dB. It achieves 75.70\% and 58.55\% phoneme accuracy respectively, over 14 percentage points better than the state-of-the-art for all noise levels.
\end{abstract}


\section{Introduction}
\label{sec:intro}

Current automatic speech recognitions (ASR) systems perform very well for clean audio but not when noise is present \cite{moreno1996speech}, which limits their use in practical situations. On top of that, the best current systems use complex neural networks which require lots of memory and computations, making implementation on embedded systems difficult. For practical embedded applications, resource-efficient systems with high noise tolerance are needed.

To improve the performance under noise, several techniques have been proposed. Some systems estimate the noise envelope from the signal and subtract this from the noisy speech (Spectral Subtraction). \cite{ganapathy2009temporal} achieved 30.1\% phoneme accuracy on the TIMIT dataset for 0 dB audio SNR, compared to 65.4\% for clean audio. In \cite{frey2001algonquin} an algorithm is introduced to detect and compensate for different noise circumstances using a time-varying probability model of the spectra of the clean speech, noise and channel distortion. As an example, their method reduces the Word Error Rate (WER) on the Wall Street Journal dataset from 28.8\% to 12.6\%, compared to a clean audio WER of 4.9\%. 
These techniques can improve ASR performance but often only under certain circumstances, and their performance still lags significantly behind clean audio performance. 
 
\par
An alternative way of tackling this problem is to use visual information by performing lipreading in addition to possibly noisy acoustic signals, a technique humans use as well.
Several authors have proposed automated lip-reading ASR systems. A common approach is to first generate feature vectors from video frames, which are then processed by a Hidden Markov Model (HMM). Lan et al \cite{lan2009comparing} take this approach and compare several feature extraction techniques, such as Active Appearance Modeling (AAM), sieve features, eigenlips and DCT-based techniques. The best system achieves 35\% word error rate (WER) on the GRID lipreading dataset.

More recently, Deep Learning methods have been applied for visual ASR. These allow learning directly from raw data and do not require expert knowledge, while showing better performance than hand-crafted feature recognizers. An often used dataset is GRID, even though it is not suited for real-life speech recognition due its very simple sentence structure and its lack of phoneme labels. In \cite{wand2016lipreading} Long-Short-Term-Memory (LSTM) neural networks were used for lipreading, achieving a 20.4\% WER on GRID, while Assael et al. \cite{lipnet_assael2016} use spatio-temporal Convolutional Neural Networks (CNN) and achieved a WER of 4.8\%.
Chung et al. \cite{WLAS2016} use a CNN-LSTM architecture for lipreading, achieving a WER of 3.0\% on GRID. They also perform audio-visual speech recognition on their own LRS dataset, combining the lipreading CNN-LSTM with an LSTM for audio recognition through an attention network. 

This work uses a similar setup, but takes the system's required resources in terms of necessary operations and memory into account. Furthermore, it analyzes the performance of the audio-visual system under noisy audio conditions.
In Section \ref{sec:audioOnly}, LSTM networks for acoustic only ASR are evaluated on the TIMIT dataset. The section discusses the performance versus resource requirements trade-off in different network architectures. Section \ref{sec:lipOnly} discusses CNNs for lipreading, which are then combined with LSTM networks in order to model temporal information \cite{WLAS2016}. Also here, a solution is optimal when it requires the least amount of resources.
In Section \ref{sec:audioVisual}, acoustic and visual networks using the least resources are combined through an attention network.

\section{Dataset and implementation details} \label{sec:implementation}

Large, varied and well-labeled audio-visual datasets are expensive to create and therefore often non-public (for example \cite{amodei2015deep}). Until recently, the largest public audio-visual dataset was GRID, which uses only very simple sentence structures, making it unsuitable to train for real-life ASR. 
Chung et al. used video from several BBC programs to generate a large-scale 'Lip Reading Sentences' (LRS) dataset \cite{WLAS2016} which was not yet released during this work.  The LRS dataset only has sentence-level labeling, in contrast to TCD-TIMIT which also contains phoneme-level labels. Though TCD-TIMIT is smaller than LRS, it still contains more data of higher quality than most other public datasets \cite{harte2015tcd}. It contains videos from 59 normal speakers and 3 professional lipspeakers. We evaluate our work on the TCD-TIMIT lipspeaker dataset for acoustic, visual and audio-visual phoneme recognition.

TCD-TIMIT label files have been created through force-alignment with the P2FA tool \cite{yuan2008speaker} and use the reduced phoneme set by Lee and Hon \cite{lee1989speaker}. For acoustic speech recognition the TIMIT dataset is used as well, with the same phoneme set.
All networks are trained using backpropagation Adam optimization with Cross-Entropy loss, using an initial learning rate of 0.01 with exponential decay of 0.5. Training is stopped when the validation loss does not decrease for 5 epochs.
All networks are implemented in Python, using the Theano \cite{theano} and Lasagne \cite{lasagne} frameworks.

\section{Acoustic Phoneme Recognition} \label{sec:audioOnly}

There are several networks that perform well for audio-only speech recognition. Hierarchical convolutional deep maxout networks \cite{toth2015} currently hold the state-of-the-art on the TIMIT dataset (83.5\% phoneme accuracy).  Graves et al. achieved 82.3\% using Deep Bidirectional LSTM networks (DBLSTM) with Connectionist Temporal Classification (CTC) \cite{graves2013_DRNNspeech}. This paper also uses DBLSTM networks as they are relatively simple.
Instead of applying CTC with beam search to find the most probable phoneme sequence, the force-aligned label files provided in the TCD-TIMIT dataset are used. For each time interval specified in the label file, the top phoneme prediction at the audio frame corresponding to the middle of that interval is selected. As there is no extra decoding step, training is faster and this allows for more direct comparison of network performance.

Deep neural networks are state-of-the-art, but the most accurate require up to hundreds of MB of storage and billions of operations per evaluation. This is too costly for embedded systems, hence this resource-accuracy trade-off should be well considered.
Each weight in an Recurrent Neural Network (RNN) requires one Multiply-Accumulate (MAC) operation, or two floating point operations (FLOP), for the processing of each audio frame.

\begin{figure}
	\centering
	\includegraphics[width=0.48\textwidth]{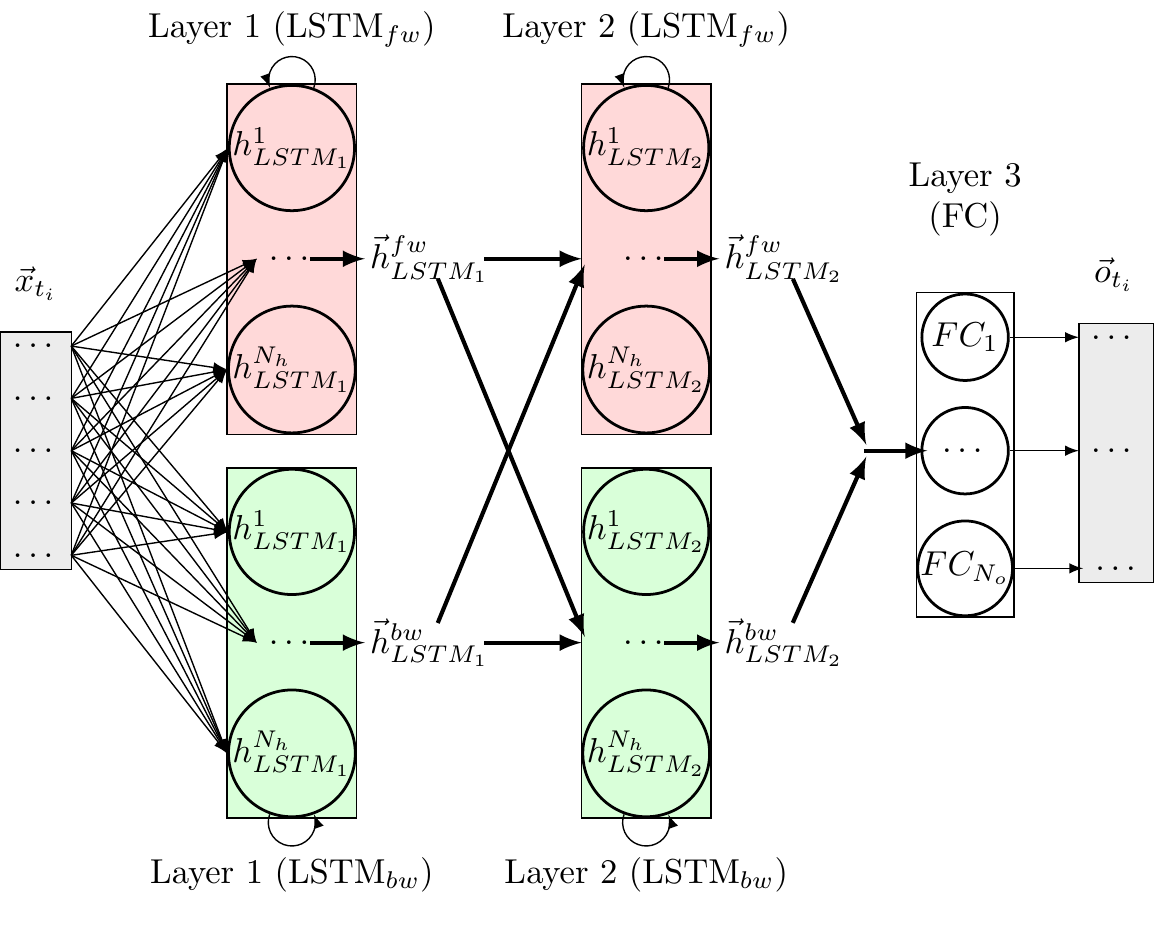}
	\caption{A network with $N_L=2$ bidirectional LSTM layers, $N_h$ LSTM units per sublayer, and one Fully Connected (FC) layer of $N_o$ neurons. At timestep $t_i$, $\vec{x}_{t_i}$ is the input vector and $\vec{o}_{t_i}$ is the network output.
		Bidirectional layers process the sequence of input vectors $\vec{x}_{t_i} ... \vec{x}_{t_L}$. The output of the forward and backward layers for each timestep is the next layer's input sequence. }
	\label{fig:DBLSTM}
\end{figure}

The audio is converted to 12 mel-frequency cepstral coefficients (MFCCs) plus the energy, with first and second derivatives, resulting in a frame of 39 features for each 'timestep' of 10 ms of audio. These are normalized using mean and variance of the training set.

RNNs process a stream of inputs, using internal states as a memory to model time dependencies. At each timestep, an vector of input features is processed and the internal states of the neurons are updated. 
In LSTMs the neurons are replaced by LSTM units, improving performance for long time-sequences \cite{hochreiter1997long}. 
In a bidirectional network each layer consists of two recurrent layers. The input stream is split in sequences of length L, the number of timesteps in an input sentence, and two recurrent layers processes this sequence in opposite directions. This allows the network to use both past and future information for its predictions, improving performance. The outputs of both layers are summed for each timestep, and passed on to subsequent layers. Fig. \ref{fig:DBLSTM} shows the network architecture of a deep bidirectional LSTM network. 
 As each layer is now replaced by two sublayers the number of neurons and weights in the network doubles. However, performance increases significantly as well \cite{graves2013_DRNNspeech}. After all recurrent layers, one feed-forward layer performs classification.

All tested network are bidirectional and the number of LSTM units across layers is kept equal, so each network is fully defined by the number of layers $N_L$ and the number of LSTM units per layer $N_h$. For simplicity, $N_h$ denotes the number of neurons in the forward LSTM layer only. 
All tested networks are first trained using the TIMIT dataset for comparison with other authors. Then they are retrained on the TCD-TIMIT dataset and their performance is compared with TCD-TIMIT baseline results \cite{harte2015tcd}.

\begin{table}
	\centering
	\small
	\renewcommand{\arraystretch}{1} 
	\caption{TIMIT Acoustic Phoneme accuracy with DBLSTMs. $N_L$ is the number of layers and $N_h$ the number of LSTM units per layer  }
	\begin{tabular}{lllll}
		$\mathbf{N_h}$ & $\mathbf{N_L}$=1 & $\mathbf{N_L}$=2 & $\mathbf{N_L}$=3 & $\mathbf{N_L}$=4 			 \\\hline
		8                                                         & 63.93   & 64.22    & 65.02    & 64.67    \\
		32                                                        & 75.99   & 76.70    & 76.74    & 77.67    \\
		64                                                        & 77.08   & 79.48    & 79.90    & 78.66    \\
		256                                                       & 78.76   & 80.98    & 80.93    & 79.42    \\
		512                                                       & 78.54   & 81.61    & 82.04    & 79.34    \\
		1024                                                      & 78.42   & 79.56    & 81.42    & 81.64   
	\end{tabular}
	\label{t:audioNetworks}
\end{table}

\begin{figure}
	\centering
	\includegraphics[width=0.45\textwidth]{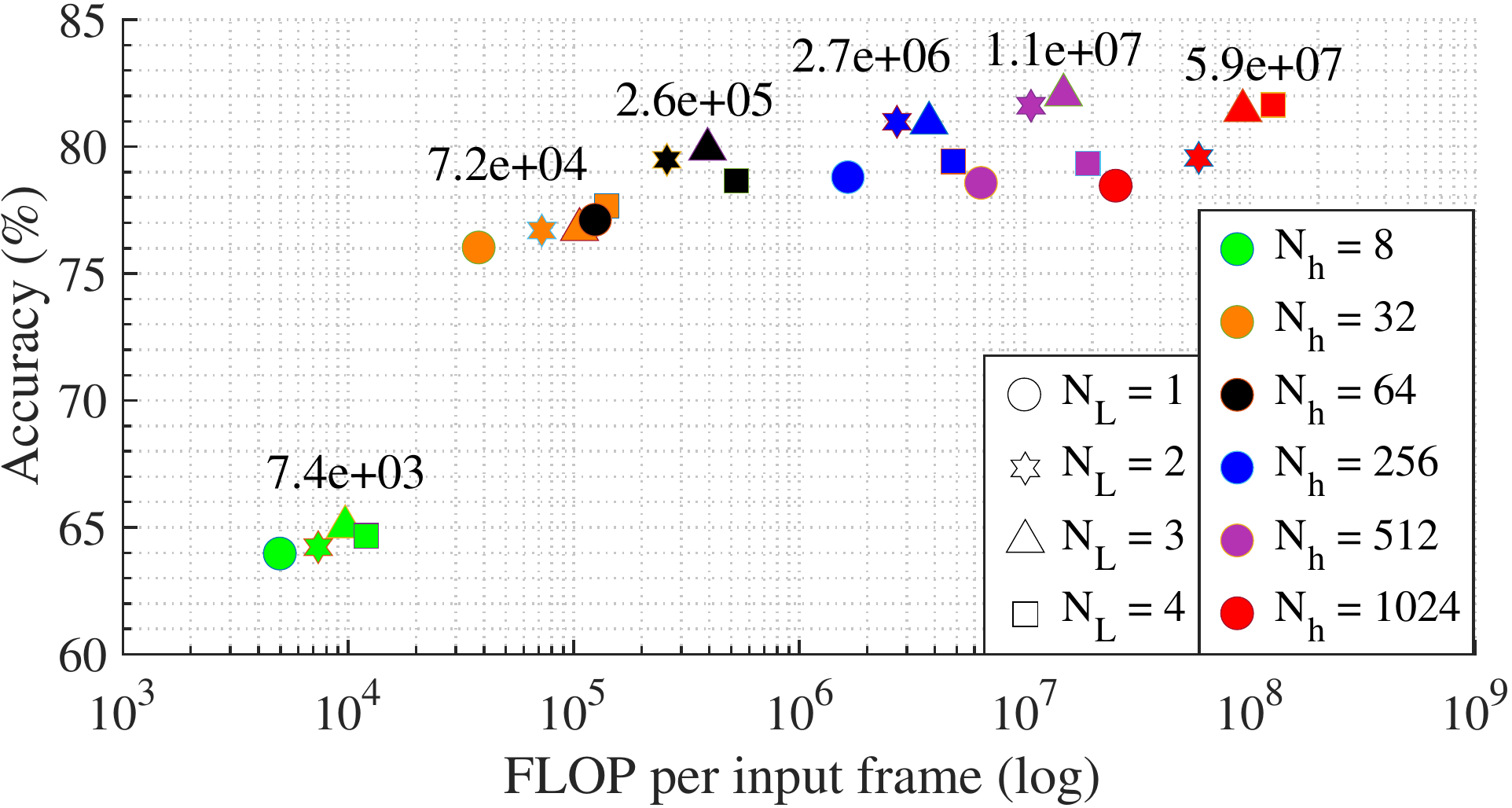}
	\caption{The trade-off between TIMIT phoneme accuracy and network size for DBLSTM networks, when $N_L$ and $N_h$ are varied. The labels denote the FLOP per audio input frame for $N_L=2$. }
	\label{fig:audioWeights}
\end{figure}

\begin{figure}[t]
	\centering
	\includegraphics[width=0.43\textwidth]{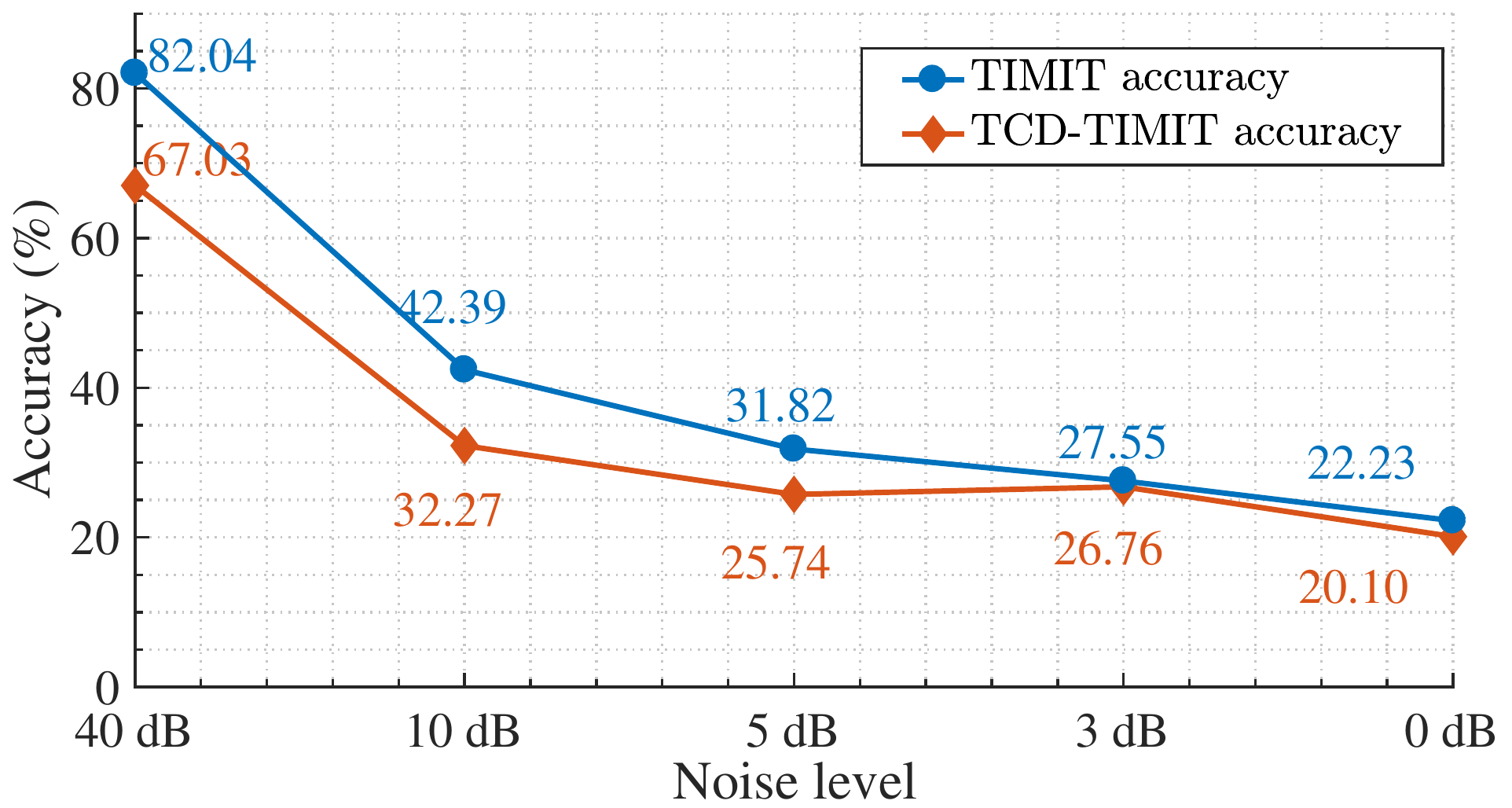}
	\caption{Acoustic phoneme classification with additive white noise ($N_L$=3 and $N_h$=512). Performance drops significantly.}
	\label{fig:audioTIMITvsTCD}
\end{figure}

Table \ref{t:audioNetworks} shows an overview of the phoneme recognition performance of several network architectures on TIMIT, while Fig. \ref{fig:audioWeights} shows the trade-off between this computational complexity and performance when the number of number of units per layer $N_h$ and the number of layers $N_L$ are varied.
The network with $N_L$=2 and $N_h$=256 offers high performance (80.98\% accuracy) at limited cost and is selected for use in the final audio-visual network (Section \ref{sec:audioVisual}). The  $N_L$=3, $N_h$=512 network achieves slightly higher performance (82.04\%) but requires six times more computations and memory. In \cite{graves2013_DRNNspeech} 82.3\% was achieved, so this is a competitive result.
The baseline on TCD-TIMIT is 65.46\%, while our highest result is 67.03\% with $N_L$=3 and $N_h$=512.  Fig. \ref{fig:audioTIMITvsTCD} shows the performance degradation of this network when white noise is added to the audio. Performance degrades to 22.23\% for TIMIT and 20.10\% for TCD-TIMIT. This confirms the need for a noise-robust audio-visual ASR solution.

\section{Visual Phoneme Recognition}\label{sec:lipOnly}

Visual ASR systems often uses visemes as class labels instead of phonemes. Visemes are visually indistinguishable features, and several phonemes can map to a single viseme. Several phoneme-to-viseme mappings have been proposed, but it is unclear whether using such a mapping is useful at all. In \cite{bear2016decoding} it was found that a word-level lipreading system using phoneme-based DNN-HMMs can perform better than a viseme-based system. Therefore this work uses phonemes as class labels. The inputs are greyscale 120x120 pixel images of the speakers' mouth area. For faster training, only the image frames corresponding to the middle of each phoneme interval in the force-aligned label file are processed.


As Convolutional Neural Networks (CNNs) have shown very good performance on many image recognition tasks, three different CNN architectures are compared: first a simple CNN architecture 'CNN$_{chung}$' \cite{WLAS2016}, second a CNN with 13 layers 'VGG$_{13}$' based on VGG$_{16}$ \cite{simonyan2014very}, and third a residual neural network with 50 layers 'ResNet$_{50}$' \cite{he2016deep}. The VGG$_{13}$ network removes the last convolutional-pooling stage from VGG$_{16}$ to retain a sufficient number of features in the CNN output. Table \ref{t:lipreading} shows the results, together with the necessary resources.
Performance of the three CNNs is relatively close. Memory and computational cost are highest for VGG$_{13}$, followed by ResNet$_{50}$, and lowest for CNN$_{chung}$, which offers decent performance at low FLOP and memory requirements. 

\begin{figure}[h]
	\centering
	\includegraphics[height=0.122\paperheight]{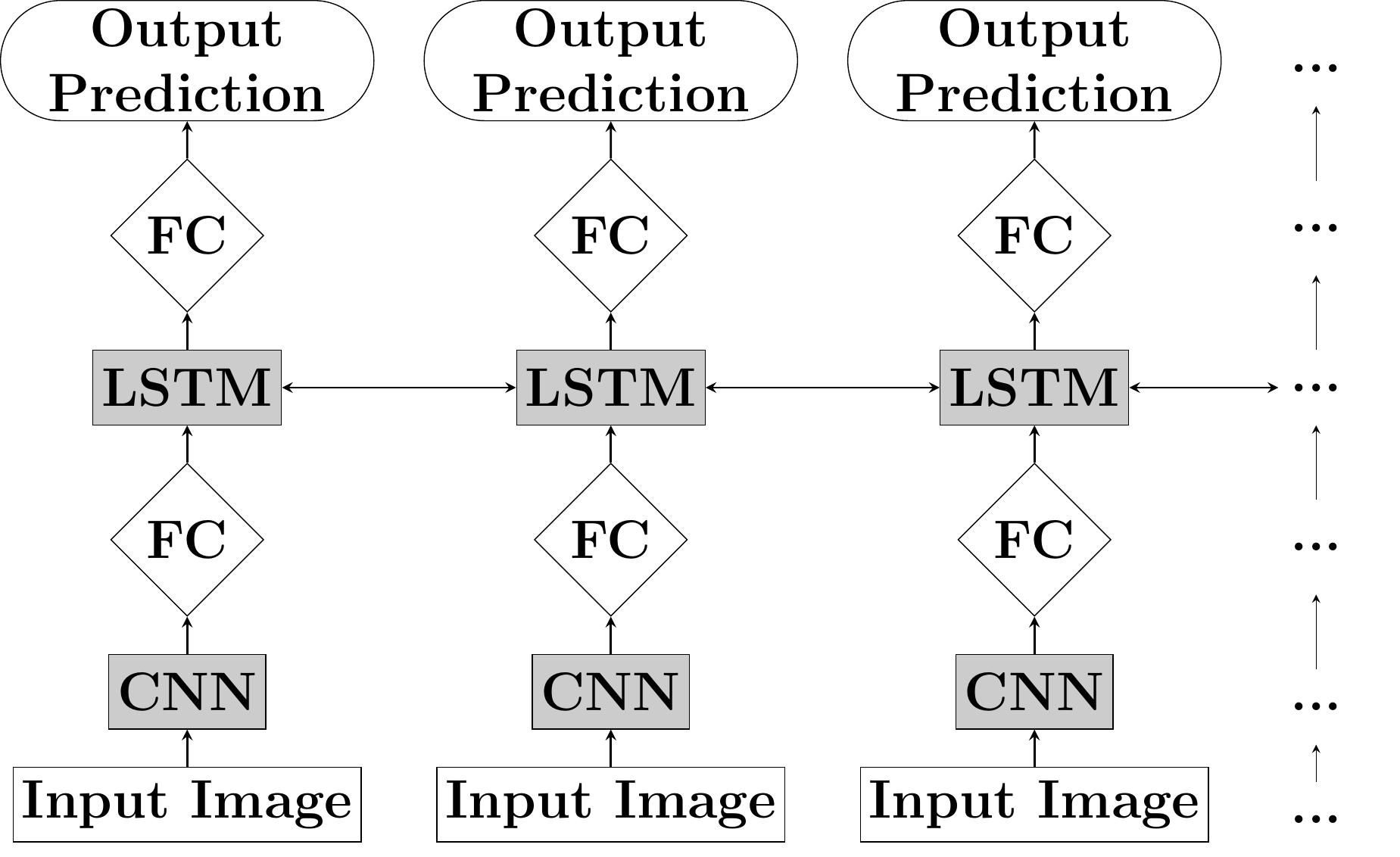}
	\caption{CNN-FC-LSTM architecture. Each column represents the network at one timestep, processing one input image. The CNN outputs for each image in the input sequence form a sequence that is processed by the (bidirectional) LSTM layers.}
	\label{fig:phonemeCNN_LSTM_PF_archPaper}
\end{figure}

\begin{figure}
	\centering
	\includegraphics[width=0.43\textwidth]{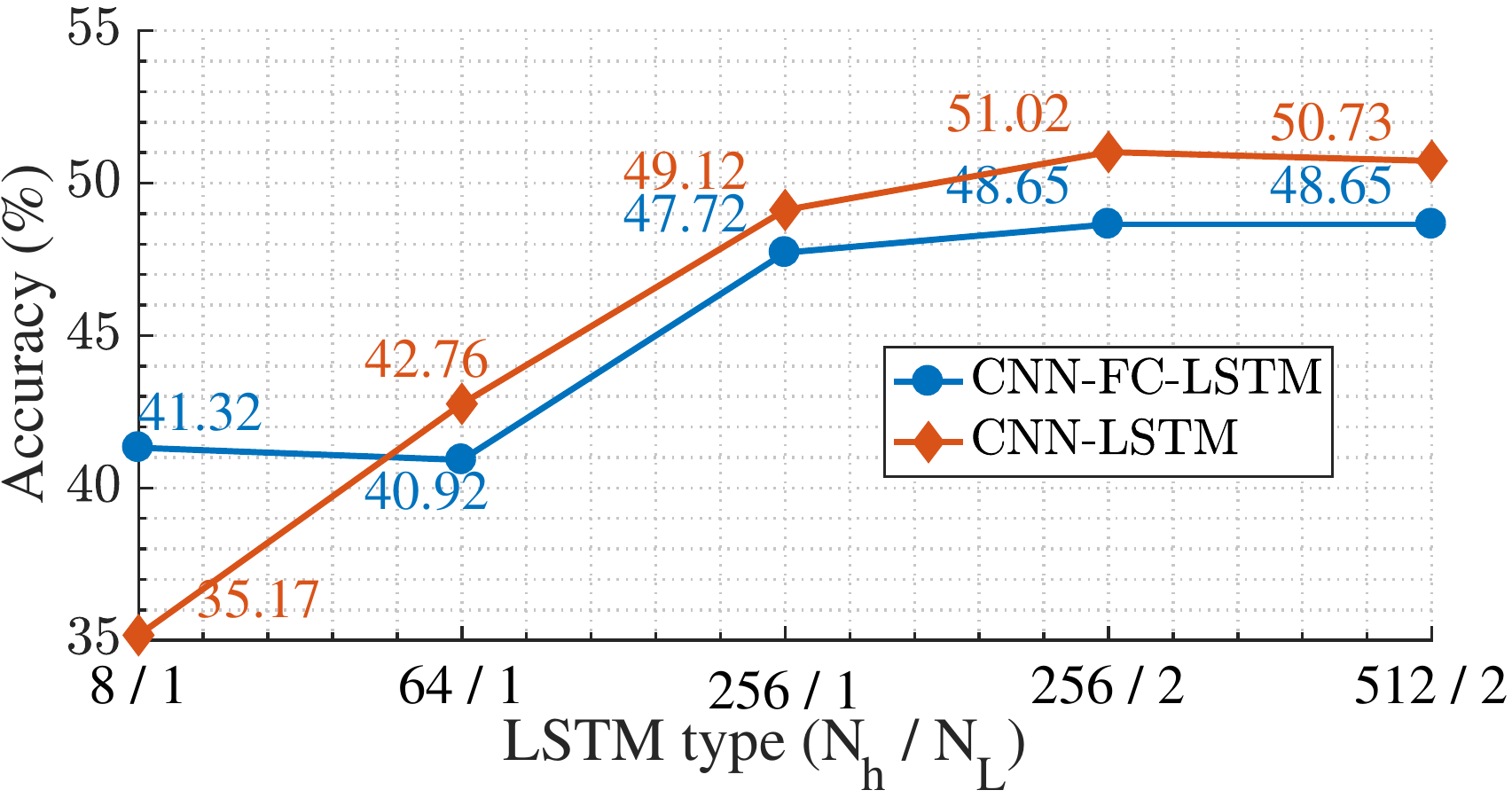}
	\caption{Visual Phoneme Classification (TCD-TIMIT 'lipspeakers'): CNN-FC-LSTM and CNN-LSTM architectures with CNN$_{chung}$.}
	\label{fig:CNN_LSTM_comparison}
\end{figure}
For speech recognition using temporal information is important, which can be achieved by adding bidirectional LSTM layers to the CNN network. The CNN output features can be directly fed to the LSTM layers (CNN-LSTM) or a Fully Connected (FC) layer can be added in between (CNN-FC-LSTM). Fig. \ref{fig:phonemeCNN_LSTM_PF_archPaper} shows the CNN-FC-LSTM architecture while Fig. \ref{fig:CNN_LSTM_comparison} compares the performance of the two architectures for several LSTM networks.
The FC layer between CNN and LSTM has 39 neurons with softmax nonlinearity, so its outputs could also be interpreted as phoneme probabilities.

An overview of the performance of our networks is shown in Table \ref{t:lipreading}. Adding LSTM layer improves performance dramatically, achieving best performance for $N_L$=2 and $N_h$=256. Replacing CNN$_{chung}$ by a residual network with 50 layers slightly improved performance (49.05\% compared to 48.65\%), at the cost of higher memory requirements.
Without the FC layer between CNN and LSTM performance increases. However, the CNN produces a large number of output features. LSTM layers are fully connected so this large number of features leads to a huge amount of weights in the LSTM. Adding a FC layer after the CNN can reduce the number of CNN output features, and thus the number of weights in the LSTM. For embedded platforms, a CNN-FC-LSTM architecture with a simple CNN like CNN$_{chung}$ is recommended.

\begin{table}
	\begin{flushleft}
		\small
		\renewcommand{\arraystretch}{1}
		\caption{Visual Phoneme and Viseme Classification (TCD-TIMIT Lipspeakers). Baseline results from \cite{harte2015tcd}. LSTM has $N_L$=2, $N_h$=256.}
		\label{t:lipreading}
		
		\begin{tabular*}{\columnwidth}{l @{\extracolsep{\fill}} c@{\extracolsep{\fill}} c@{\extracolsep{\fill}} c@{\extracolsep{\fill}} c@{\extracolsep{\fill}}}
			\textbf{Network}		 	& \textbf{FLOP / image} 	& \textbf{Size (MB)}  & \textbf{Phn.} 	& \textbf{Vis.} \\\hline
			CNN$_{chung}$        	 	& $1.05 * 10^9 $   	& 27.39               & 39.90    		&	55.55\\
			ResNet$_{50}$    	 		& $1.18 * 10^9 $  	& 90.21               & 42.55    		&	54.42\\
			VGG$_{13}$      			& $4.14 * 10^9 $   	& 485.77              & 41.91    		& 	56.31\\
			CNN$_{chung}$-FC-LSTM 		& $1.05 * 10^9 $   	& 32.59               & 48.65   		&	63.83\\
			ResNet$_{50}$-FC-LSTM		& $1.18 * 10^9 $	& 95.39		     	  & 49.05			&	-\\
			CNN$_{chung}$-LSTM    		& $1.08 * 10^9 $   	& 126.7               & 51.02    		&	68.46\\
			Baseline (HMM)   			& -   				& -            		  & -    			&	57.85\\
		\end{tabular*}
	\end{flushleft}
\end{table}

Baseline visual-only viseme classification accuracy for TCD-TIMIT is 57.85\% using HMM-based classifiers on a DCT of the mouth region. To compare our results with the baseline, the data labels are first mapped to visemes using the phoneme-to-viseme mapping developed by Neti et al. \cite{neti2000audio}, and the networks are retrained for viseme classification. Table \ref{t:lipreading} shows the results. The viseme classification scores are much higher than the phoneme classification scores as several phonemes map to a single viseme.
The CNN-only networks perform worse than the TCD-TIMIT baseline, but the combination of CNN and LSTM performs significantly better, achieving an improvement of over 10\% over the HMM-based viseme baseline.

\section{Audio-visual Phoneme Recognition}\label{sec:audioVisual}

\begin{figure}
	\centering
	\includegraphics[width=0.50\textwidth]{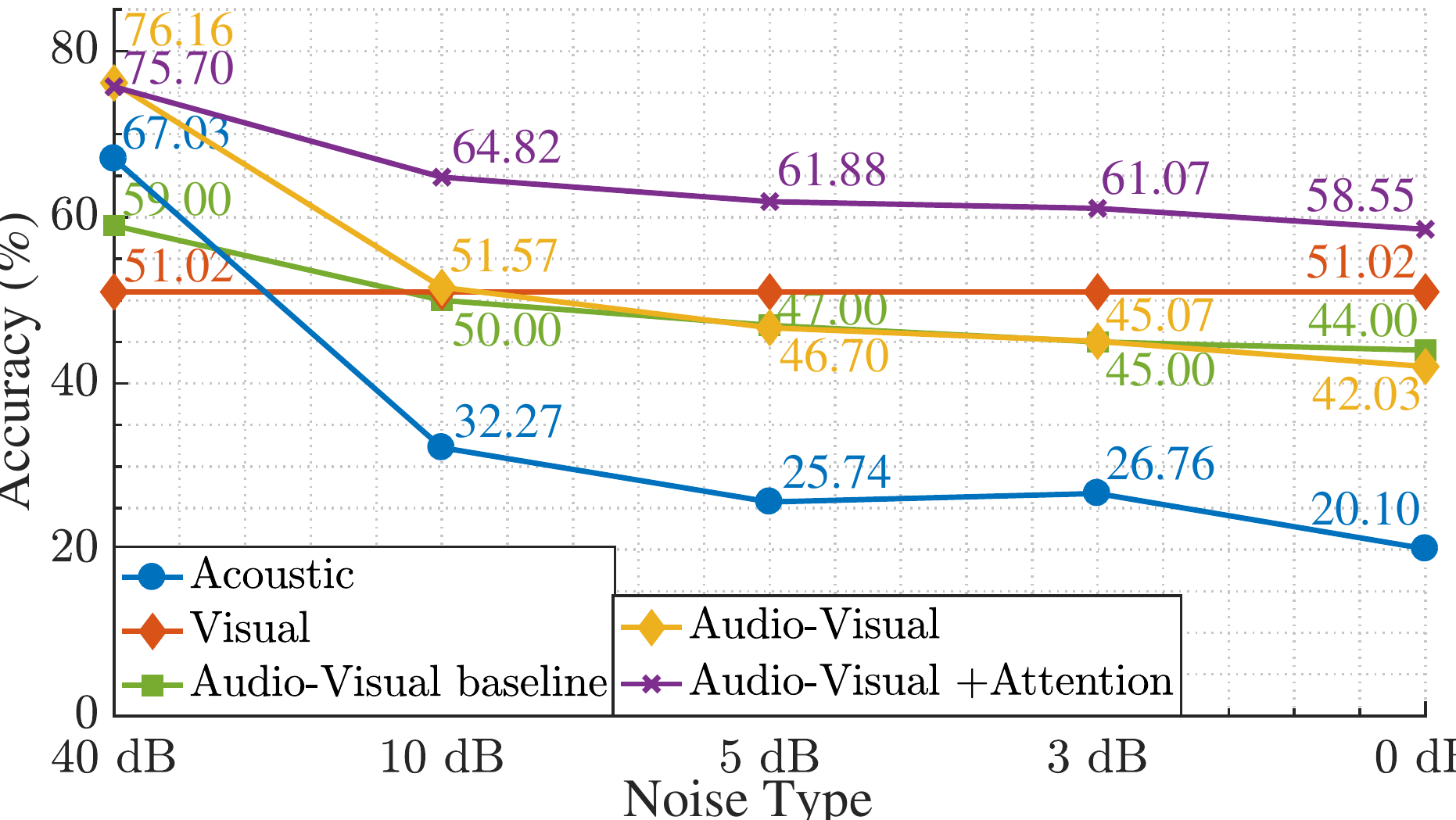}
	\caption{Phoneme Classification: acoustic, visual and audio-visual networks and audio-visual baseline (TCD-TIMIT 'lipspeakers')}
	\label{fig:multimodalNoise}
\end{figure}

Acoustic and visual networks can be combined by feeding their concatenated outputs at each timestep, as input to the fully connected classification layers. Three layers of 512 neurons each are used. Section \ref{sec:audioOnly} showed that for the acoustic network, an LSTM network with 2 layers and 256 LSTM units per layer performed well at fairly low memory and computation requirements, so this architecture is used for the acoustic model and and after CNN$_{chung}$ in the visual network. The acoustic and visual networks are first trained separately, then the combined network is built using the pretrained weights, and the entire network is trained with backpropagation.

\begin{figure}[t]
	\centering
	\includegraphics[width=0.45\textwidth]{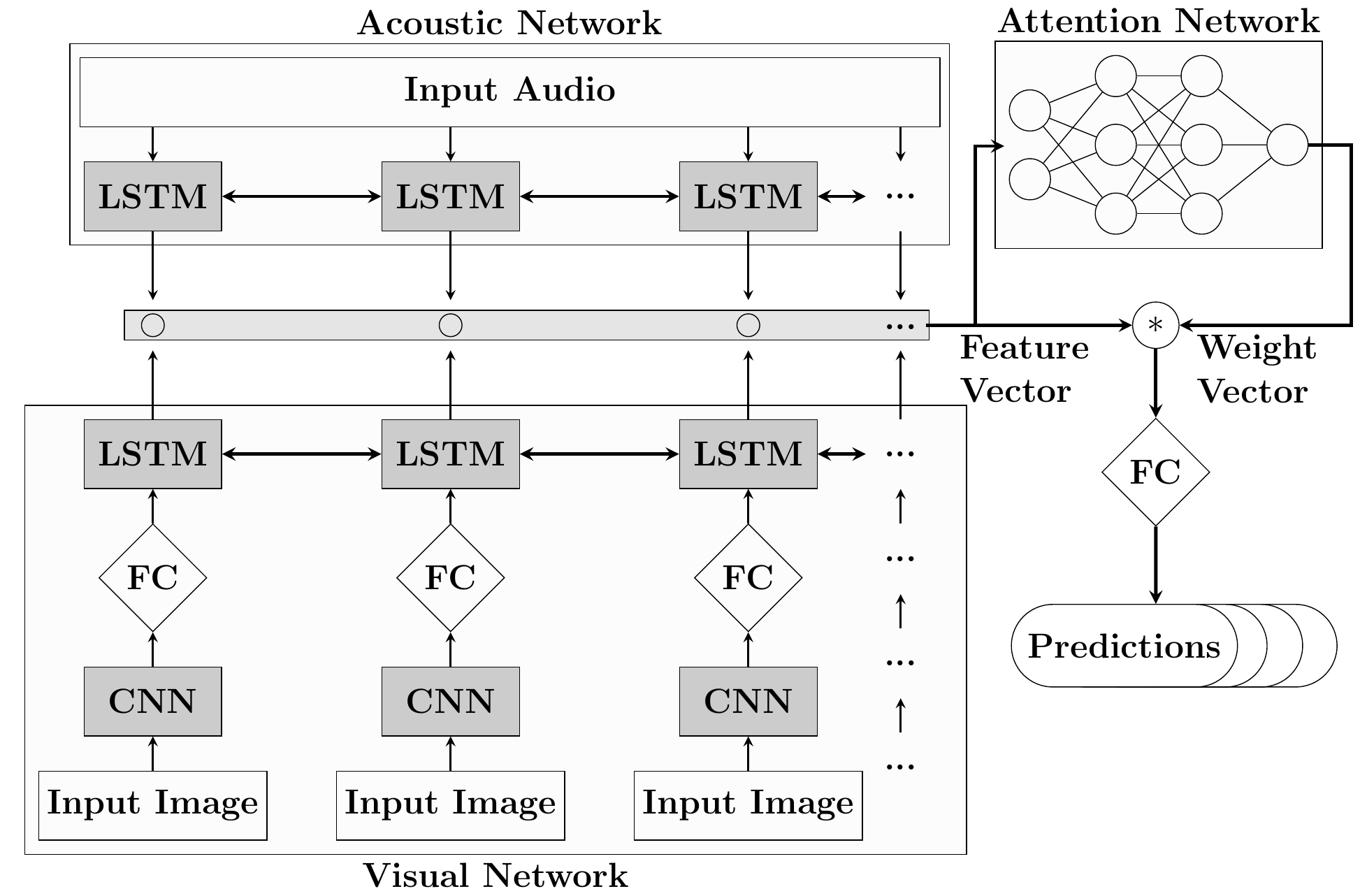}
	\caption{The architecture of an audio-visual network with attention. }
	\label{fig:multimodalArch}
\end{figure}

Fig. \ref{fig:multimodalNoise} shows the performance of acoustic and visual networks, of the audio-visual network with and without attention, and of the TCD-TIMIT audio-visual baseline \cite{harte2015tcd}. To our knowledge, this baseline is currently the state-of-the-art (SotA) for audio-visual phoneme recognition on the TCD-TIMIT lipspeaker dataset.
Fig. \ref{fig:multimodalArch} shows the full final network architecture with an added attention mechanism. Without attention, the audio-visual feature vector directly goes to the FC network. Without attention, the audio-visual network achieves 76.19\% for clean audio and 42.03\% for 0dB audio SNR. This is significantly better than either acoustic or visual networks for clean audio. However, performance is worse than for the visual-only network. This is because the fully connected layers attribute a fixed relative importance to acoustic and visual features so the network cannot adapt to changes in audio quality. Intuitively, when audio quality is low visual features should be valued higher, while acoustic features should be valued higher when audio quality is high. 

An attention system can intelligently combine the acoustic and visual features, by estimating how certain each subnetwork is of its predictions and varying the relative weights of the features correspondingly. This attention network (AN) can be implemented as a feed-forward network as shown in Fig. \ref{fig:multimodalArch}, where the inputs are the outputs of the acoustic and visual networks. The AN consists of three layers of 512 neurons each. 
The network is first trained with clean audio, then retrained with varying levels of additive white noise to enable the AN to learn the optimal relative audio-visual weighting in each noise condition.

Table \ref{t:audio-visual} illustrates that at 0dB audio SNR, the audio-visual network with attention performs much better than without attention. It is also better than visual-only networks and audio-only networks at any noise level. Fig. \ref{fig:multimodalNoise} shows the performance for different noise levels.
The audio-visual network with attention achieves 14\% absolute improvement over the TCD-TIMIT audio-visual baseline across all noise levels. Table \ref{t:audio-visual} compares the best solutions in terms of performance and resource requirements. The baseline resource costs are not given in \cite{harte2015tcd}, but model size is estimated from \cite{yuan2006speech}.
The results show that combining acoustic and visual data significantly improves ASR performance, particularly in noisy conditions. The attention mechanism is crucial to achieve high performance across noise levels. 

\begin{table}
		\centering
		\small
		\renewcommand{\arraystretch}{1.1}
		\caption{Comparison of top performing networks. FLOP per second (100 audio frames/s, 30 image frames/s)}
		\label{t:audio-visual}
		\begin{tabular*}{\columnwidth}{l @{\extracolsep{\fill}} c@{\extracolsep{\fill}} c@{\extracolsep{\fill}} c@{\extracolsep{\fill}} c@{\extracolsep{\fill}}}
			\textbf{Network}	& \textbf{FLOPS }		& \textbf{Size (MB)}		& \textbf{40dB}			&  \textbf{0 dB} \\\hline
			Acoustic-only    	& 100x $1.49 \cdot 10^7 $	& 44.5		& 67.03 	  	& 20.10         \\
			Visual-only        	& 30x $1.08 \cdot 10^9 $	& 132.9		& 51.02    		& 51.02         \\
			Audio-Visual        & 30x $1.12 \cdot 10^9 $	& 134.9 	& 76.19    		& 42.03         \\
			Audio-Visual + Att. & 30x $1.12 \cdot 10^9 $	& 137.4		& 75.70    		& 58.55         \\
			Audio-Visual SotA \cite{harte2015tcd} 	& -		& $\approx$1.13		& 59    		& 44           
		\end{tabular*}
\end{table}

\section{Conclusion} \label{sec:conclusion}
This work discusses the resource-aware design of a noise-resistant audio-visual phoneme recognition system achieving state-of-the art results on the TCD-TIMIT lipspeaker dataset. For each of the building blocks of this audio-visual system, the trade-off between performance and required resources is analyzed.

First, LSTM-based audio networks are shown to perform well. On TCD-TIMIT the highest achieved performance is 67.03\% phoneme accuracy, compared to the baseline performance of 65.47\%. 

Second, CNN-based neural networks are evaluated for lipreading. CNN-LSTM architectures perform much better than CNN-only networks, achieving 68.46\% viseme accuracy compared to 56.31\% respectively. The TCD-TIMIT baseline architecture achieves 57.85\% viseme accuracy, which is over 10 percentage points less. On phonemes, highest accuracy is 42.55\% using a CNN-only network and 51.02\% for a CNN-LSTM network.

Finally, acoustic and visual networks are combined through an attention mechanism to increase recognition performance in noisy conditions. This audio-visual network achieves 75.70\% phoneme accuracy for clean audio and 58.55\% at 0dB audio SNR, which is an up to 16\% absolute improvement on the audio-visual TCD-TIMIT baseline that achieves 59\% and 44\% respectively. 
The best system requires a larger budget than the baseline HMM-based system at $30 \cdot 10^9$ FLOPS and 137.4 MB of model storage, but the performance is significantly higher both for clean and noisy audio.

\vfill\pagebreak

\bibliographystyle{ieeetr}
\bibliography{paper}

\begin{thebibliography}{10}

\bibitem{moreno1996speech}
P.~J. Moreno, ``Speech recognition in noisy environments,'' 1996.

\bibitem{ganapathy2009temporal}
S.~Ganapathy, S.~Thomas, and H.~Hermansky, ``Temporal envelope subtraction for
  robust speech recognition using modulation spectrum,'' in {\em Automatic
  Speech Recognition \& Understanding, 2009. ASRU 2009. IEEE Workshop on},
  pp.~164--169, IEEE, 2009.

\bibitem{frey2001algonquin}
B.~J. Frey, L.~Deng, A.~Acero, and T.~T. Kristjansson, ``Algonquin: iterating
  laplace's method to remove multiple types of acoustic distortion for robust
  speech recognition.,'' in {\em INTERSPEECH}, pp.~901--904, 2001.

\bibitem{lan2009comparing}
Y.~Lan, R.~Harvey, B.~Theobald, E.-J. Ong, and R.~Bowden, ``Comparing visual
  features for lipreading,'' in {\em International Conference on
  Auditory-Visual Speech Processing 2009}, pp.~102--106, 2009.

\bibitem{wand2016lipreading}
M.~Wand, J.~Koutn{\'\i}k, and J.~Schmidhuber, ``Lipreading with long short-term
  memory,'' in {\em Acoustics, Speech and Signal Processing (ICASSP), 2016 IEEE
  International Conference on}, pp.~6115--6119, IEEE, 2016.

\bibitem{lipnet_assael2016}
Y.~M. Assael, B.~Shillingford, S.~Whiteson, and N.~de~Freitas, ``Lipnet:
  Sentence-level lipreading,'' {\em arXiv preprint arXiv:1611.01599}, 2016.

\bibitem{WLAS2016}
J.~S. Chung, A.~Senior, O.~Vinyals, and A.~Zisserman, ``Lip reading sentences
  in the wild,'' {\em arXiv preprint arXiv:1611.05358}, 2016.

\bibitem{amodei2015deep}
D.~Amodei, R.~Anubhai, E.~Battenberg, C.~Case, J.~Casper, B.~Catanzaro,
  J.~Chen, M.~Chrzanowski, A.~Coates, G.~Diamos, {\em et~al.}, ``Deep speech 2:
  End-to-end speech recognition in english and mandarin,'' {\em arXiv preprint
  arXiv:1512.02595}, 2015.

\bibitem{harte2015tcd}
N.~Harte and E.~Gillen, ``Tcd-timit: An audio-visual corpus of continuous
  speech,'' {\em IEEE Transactions on Multimedia}, vol.~17, no.~5,
  pp.~603--615, 2015.

\bibitem{yuan2008speaker}
J.~Yuan and M.~Liberman, ``Speaker identification on the scotus corpus,'' {\em
  Journal of the Acoustical Society of America}, vol.~123, no.~5, p.~3878,
  2008.

\bibitem{lee1989speaker}
K.-F. Lee and H.-W. Hon, ``Speaker-independent phone recognition using hidden
  markov models,'' {\em IEEE Transactions on Acoustics, Speech, and Signal
  Processing}, vol.~37, no.~11, pp.~1641--1648, 1989.

\bibitem{theano}
{Theano Development Team}, ``{Theano: A {Python} framework for fast computation
  of mathematical expressions},'' {\em arXiv e-prints}, vol.~abs/1605.02688,
  May 2016.

\bibitem{lasagne}
S.~Dieleman, J.~Schlüter, C.~Raffel, E.~Olson, S.~K. Sønderby, D.~Nouri, {\em
  et~al.}, ``Lasagne: First release.,'' Aug. 2015.

\bibitem{toth2015}
L.~T{\'o}th, ``Phone recognition with hierarchical convolutional deep maxout
  networks,'' {\em EURASIP Journal on Audio, Speech, and Music Processing},
  vol.~2015, p.~25, Sep 2015.

\bibitem{graves2013_DRNNspeech}
A.~Graves, A.-r. Mohamed, and G.~Hinton, ``Speech recognition with deep
  recurrent neural networks,'' in {\em Acoustics, speech and signal processing
  (icassp), 2013 ieee international conference on}, pp.~6645--6649, IEEE, 2013.

\bibitem{hochreiter1997long}
S.~Hochreiter and J.~Schmidhuber, ``Long short-term memory,'' {\em Neural
  computation}, vol.~9, no.~8, pp.~1735--1780, 1997.

\bibitem{bear2016decoding}
H.~L. Bear and R.~Harvey, ``Decoding visemes: Improving machine lip-reading,''
  in {\em Acoustics, Speech and Signal Processing (ICASSP), 2016 IEEE
  International Conference on}, pp.~2009--2013, IEEE, 2016.

\bibitem{simonyan2014very}
K.~Simonyan and A.~Zisserman, ``Very deep convolutional networks for
  large-scale image recognition,'' {\em arXiv preprint arXiv:1409.1556}, 2014.

\bibitem{he2016deep}
K.~He, X.~Zhang, S.~Ren, and J.~Sun, ``Deep residual learning for image
  recognition,'' in {\em Proceedings of the IEEE Conference on Computer Vision
  and Pattern Recognition}, pp.~770--778, 2016.

\bibitem{neti2000audio}
C.~Neti, G.~Potamianos, J.~Luettin, I.~Matthews, H.~Glotin, D.~Vergyri,
  J.~Sison, and A.~Mashari, ``Audio visual speech recognition,'' tech. rep.,
  IDIAP, 2000.

\bibitem{yuan2006speech}
M.~Yuan, T.~Lee, P.~Ching, and Y.~Zhu, ``Speech recognition on dsp: issues on
  computational efficiency and performance analysis,'' {\em Microprocessors and
  Microsystems}, vol.~30, no.~3, pp.~155--164, 2006.

\end{thebibliography}

\end{document}